\documentclass[journal]{IEEEtran}

\ifCLASSINFOpdf
\else
\fi
\usepackage{subcaption}
\usepackage{graphicx}
\usepackage{booktabs}
\usepackage{makecell}
\usepackage{multirow}
\usepackage{pifont}
\usepackage{amsmath}
\usepackage{url}
\usepackage{mathtools}
\usepackage{amssymb}
\usepackage{amsfonts}
\usepackage{xcolor}
\usepackage{colortbl}
\usepackage{algorithm}
\usepackage{algorithmic}
\begin{document}

\title{OT-Drive: Out-of-Distribution Off-Road Traversable Area Segmentation via Optimal Transport}

\author{Zhihua~Zhao,~%
        Guoqiang~Li$^{*}$,~%
        Chen~Min,~%
        and~Kangping~Lu%
\thanks{This work has been submitted to the IEEE for possible publication. Copyright may be transferred without notice, after which this version may no longer be accessible.}%
\thanks{$^{*}$Corresponding author: Guoqiang Li (guoqiangli@bit.edu.cn).}%
\thanks{Zhihua Zhao and Guoqiang Li are with the School of Mechanical Engineering, Beijing Institute of Technology, Beijing 100081, China (e-mail: zhihuazhao@bit.edu.cn; guoqiangli@bit.edu.cn).}%
\thanks{Chen Min is with the Institute of Computing Technology, Chinese Academy of Sciences, Beijing 100190, China (e-mail: mincheng@ict.ac.cn).}%
\thanks{Kangping Lu is with Shandong Pengxiang Automobile Co., Ltd (e-mail: pengxiang\_lu@163.com).}%
}

\markboth{}%
{}
\maketitle
\begin{abstract}
Reliable traversable area segmentation in unstructured environments is critical for planning and decision-making in autonomous driving. 
However, existing data-driven approaches often suffer from degraded segmentation performance in out-of-distribution (OOD) scenarios, consequently impairing downstream driving tasks. 
To address this issue, we propose OT-Drive, an Optimal Transport--driven multi-modal fusion framework. 
The proposed method formulates RGB and surface normal fusion as a distribution transport problem. 
Specifically, we design a novel Scene Anchor Generator (SAG) to decompose scene information into the joint distribution of weather, time-of-day, and road type, thereby constructing semantic anchors that can generalize to unseen scenarios. 
Subsequently, we design an innovative Optimal Transport-based 
multi-modal fusion module (OT Fusion) to transport RGB and 
surface normal features onto the manifold defined by the 
semantic anchors, enabling robust traversable area segmentation 
under OOD scenarios.
Experimental results demonstrate that our method achieves 95.16\% mIoU on ORFD OOD scenarios, outperforming prior methods by 6.35\%, and 89.79\% mIoU on cross-dataset transfer tasks, surpassing baselines by 13.99\%.
These results indicate that the proposed model can attain strong OOD generalization with only limited training data, substantially enhancing its practicality and efficiency for real-world deployment.
\end{abstract}

\begin{IEEEkeywords}
traversable area segmentation, out-of-distribution, multi-modal fusion, freespace detection
\end{IEEEkeywords}

\IEEEpeerreviewmaketitle

\section{Introduction}

\IEEEPARstart{A}{utonomous} driving technology has achieved remarkable success in structured urban environments. 
However, deploying autonomous driving systems in unstructured environments---such as farm roads~\cite{philippeCollisionawareTraversabilityAnalysis2025}, rural areas~\cite{xuTransformerbasedTraversabilityAnalysis2024}, and mining zones~\cite{zhengPassableAreaSegmentation2024}---remains a significant challenge. 
Unlike urban roads, unstructured scenes lack clear boundaries, standardized traffic signs, and well-defined lane markings~\cite{minAutonomousDrivingUnstructured2024}. 
In such scenarios, reliable traversable area segmentation serves as a fundamental prerequisite, providing safety-critical boundary constraints for motion planning and control~\cite{chengSequentialSemanticSegmentation2022}.

\begin{figure}[tbp]
    \centering
    \includegraphics[width=\linewidth]{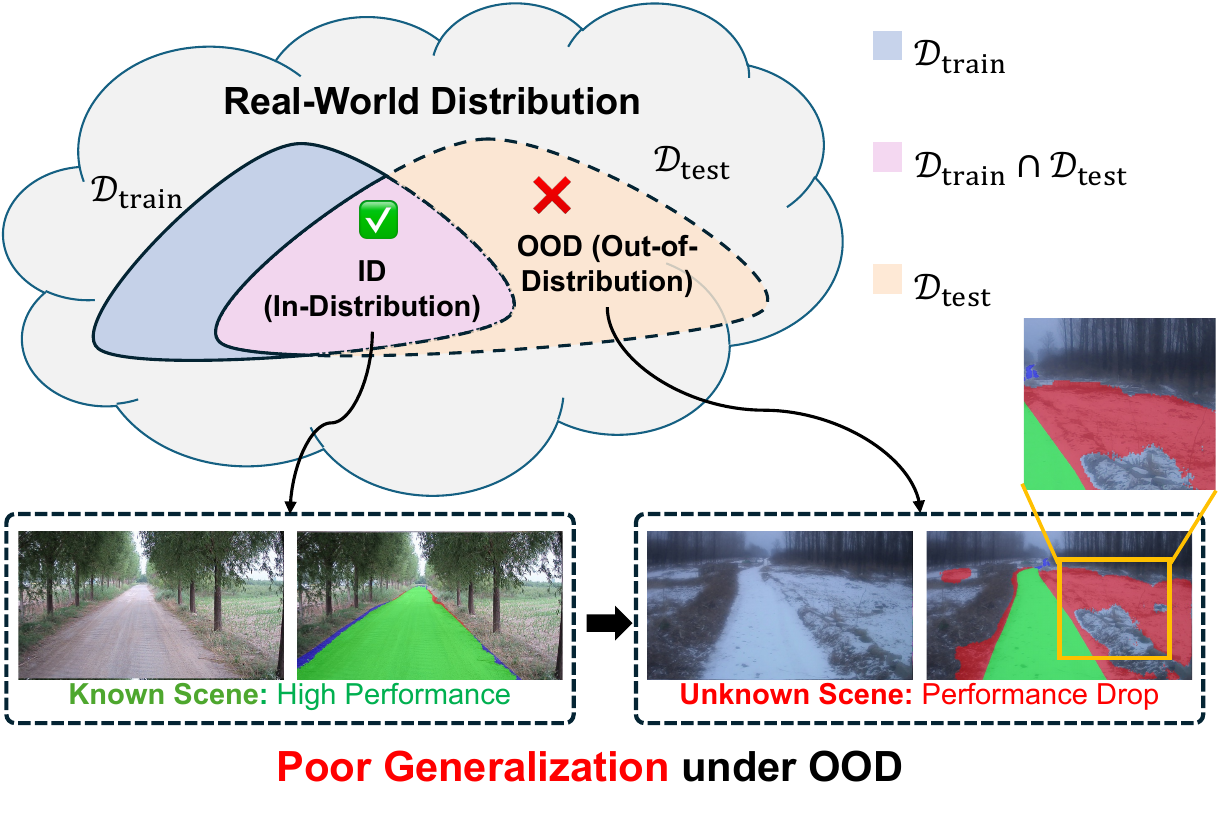}
    \caption{Existing segmentation methods exhibit poor OOD generalization. Red and green regions indicate false and true traversable areas, respectively.}
    \label{fig:problem}
\end{figure}

Capitalizing on the complementary strengths of visual textures and geometric structures, multi-modal semantic segmentation has become the prevailing paradigm for traversable area segmentation \cite{fengAdaptiveMaskFusionNetwork2023,yeM2F2netMultimodalFeature2023,minORFDDatasetBenchmark2022,kimUFOUncertaintyawareLiDARimage2024,liRoadFormerDuplexTransformer2024}. 
Under this paradigm, heterogeneous modalities---typically RGB image and surface normal---are jointly processed to predict a dense binary label for each pixel, indicating whether it is traversable or non-traversable. 
However, existing multi-modal fusion approaches often overfit to specific scene characteristics and implicitly assume that heterogeneous modalities can be directly aligned in a shared feature space. 
As a result, these methods may perform well in in-distribution (ID) scenarios that closely resemble the training data but suffer from increased false detections under OOD scenarios, as illustrated in Fig.~\ref{fig:problem}. 
Such failures can lead to catastrophic consequences, including planning trajectories into hazardous terrain (e.g., ditches or obstacles) or triggering abrupt emergency stops~\cite{fengMAFNetSegmentationRoad2022}.

To address these challenges, we propose OT-Drive, an Optimal Transport~\cite{kantorovitchTranslocationMasses1958}-driven multi-modal fusion framework for improving the generalization of traversable area segmentation in unstructured environments. 
Our core insight is to formulate RGB and surface normal fusion as a scene-conditioned distribution transport problem. 
Specifically, we first present a Scene Anchor Generator (SAG) that decomposes complex scene variations into disentangled environmental attributes---namely weather, time-of-day, and road type. 
Based on this explicit disentanglement, SAG constructs scene anchors that can generalize to novel, unseen combinations of environmental conditions. 
Furthermore, we propose an Optimal Transport-based Fusion (OT Fusion) module that, rather than forcing heterogeneous modalities into a shared feature space, seeks an optimal transport plan to align image features and surface normal features onto the manifold defined by the scene anchors. 
This mechanism achieves distribution-level optimal fusion that enables robust cross-modal alignment. Consequently, OT-Drive ensures reliable traversable area segmentation in OOD scenarios.

The \textbf{main contributions} of this work are summarized as follows:
\begin{itemize}
    \item We propose OT-Drive, a novel multi-modal fusion framework that models RGB and surface normal fusion via optimal transport, achieving robust OOD generalization for traversable area segmentation. To the best of the authors' knowledge, this is the first work to apply optimal transport to enhance the robustness of heterogeneous modality fusion under distribution shift.
    \item We design a novel Scene Anchor Generator (SAG) that disentangles scene variations into independent environmental attributes, constructing scene anchors that generalize to unseen environmental combinations.
    \item We develop an innovative Optimal Transport-based Fusion (OT Fusion) module that aligns heterogeneous modality features onto the manifold defined by scene anchors via optimal transport. 
	By minimizing the global transportation cost, this module achieves distribution-level optimal fusion that is highly robust to OOD scenarios.
    \item Extensive experiments on multiple off-road datasets and the challenging cross-dataset transfer benchmarks validate the proposed framework, demonstrating state-of-the-art OOD generalization and confirming its practicality for real-world deployment.
\end{itemize}

\section{Related Work}

\subsection{Single-Modal Traversable Area Segmentation}
Early off-road traversable area segmentation primarily relied on geometric modeling from point clouds. Pioneering works employed explicit geometric constraints, such as Markov Random Fields (MRF) for ground height estimation~\cite{jimenezGroundSegmentationAlgorithm2021} and concentric zone-based strategies for robust ground fitting, as seen in the Patchwork series~\cite{limPatchworkConcentricZonebased2021,leePatchworkFastRobust2022}. 
Transitioning to data-driven approaches, end-to-end neural networks have been widely adopted to directly predict traversability from point clouds~\cite{shabanSemanticTerrainClassification2022,seoScateScalableFramework2023}. 
More recently, hybrid architectures that project 3D point clouds onto 2D planes have enabled real-time modeling~\cite{wangSwinURNetHybridTransformerCNN2024}, while self-supervised contrastive learning has been explored to resolve label ambiguity~\cite{xuecontrastiveLabelDisambiguation2025}.
While point cloud-based methods provide strong geometric priors, they often lack the semantic understanding to differentiate structurally similar terrains.
Conversely, vision-based methods utilize grouped attention mechanisms to analyze terrain difficulty and texture~\cite{guanGanavEfficientTerrain2022,sunPassableAreaDetection2023}.
To overcome the limited field of view (FOV) of monocular cameras, recent approaches have extended perception capabilities via multi-view inputs and cross-view coordinate mappings~\cite{mengTerrainNetVisualModeling2023,chungPixelElevationLearning2024}.

In summary, single-modal approaches inherently face trade-offs between geometric accuracy and semantic richness, limiting their reliability across diverse environmental conditions.

\subsection{Multi-Modal Fusion for Traversable Area Segmentation}
To capitalize on the complementary strengths of geometry and semantics, multi-modal fusion methods integrate heterogeneous sensor data, such as camera-LiDAR~\cite{kimUFOUncertaintyawareLiDARimage2024} or hyperspectral-radar combinations~\cite{philippeCollisionawareTraversabilityAnalysis2025}. 
Given the sparsity and irregularity of 3D point clouds, a prevalent strategy involves projecting them into camera-view surface normal maps to extract dense geometric features~\cite{fanSNERoadSegIncorporatingSurface2020, wangSNERoadSegRethinkingDepthNormal2021}. 
Building on this representation, researchers have integrated surface normal with RGB image using advanced architectures---ranging from Vision Transformers~\cite{minORFDDatasetBenchmark2022} to Swin Transformers~\cite{yanFsnswinNetworkFreespace2024}---to enhance feature extraction and alignment. 
Other approaches focus on refining boundary details by employing edge decoding modules~\cite{yeM2F2netMultimodalFeature2023} or duplex transformers~\cite{liRoadFormerDuplexTransformer2024}. 
Furthermore, to mitigate the impact of sensor noise and calibration errors, recent works have developed adaptive weighting and noise-aware intermediate fusion mechanisms~\cite{lvNoiseawareIntermediaryFusion2024,fengAdaptiveMaskFusionNetwork2023}.

However, most existing methods rely on pixel-level or token-level alignment, which often fails to capture distribution shifts in OOD scenarios.

\subsection{Domain Generalization for Semantic Segmentation}

Domain Generalization (DG) aims to mitigate performance degradation caused by distribution discrepancies between training and test domains.
Early research primarily addressed this challenge through data augmentation and domain-invariant representation learning, exploiting texture statistics from ImageNet pre-training to enhance generalization \cite{kimTextureLearningDomain2023,hwangHowRelieveDistribution2025}.
Recently, leveraging textual semantics from Vision-Language Models (VLMs) to guide visual feature alignment has become a standard practice for enhancing generalization.
One line of methods constructs diverse pseudo-target domains through text prompts or generative models to enhance robustness against environmental variations~\cite{fahesSimpleRecipeLanguageguided2024,niemeijerGeneralizationAdaptationDiffusionbased2024}. 
Another line directly incorporates text embeddings into segmentation frameworks, such as serving as decoder queries~\cite{pakTextualQuerydrivenMask2025} or semantic prototypes~\cite{zhangMambaBridgeWhere2025}, significantly improving generalization under extreme styles or OOD scenarios. 
Furthermore, recent studies explore generating OOD prompts with varying semantic distances via WordNet~\cite{songLeveragingTextDrivenSemantic2025}, optimizing cross-modal alignment~\cite{yueCrossmodalDomainGeneralization2024}, and employing domain-aware prompt learning~\cite{jeonExploitingDomainProperties2025} to alleviate semantic misalignment.

Despite significant progress in single-modal DG, existing works rarely address the distribution shifts inherent in heterogeneous multi-sensor fusion frameworks. This remains an open challenge in complex perception systems such as autonomous driving.

\section{Method}
\begin{figure*}[htbp]
	\centering
	\includegraphics[width=\linewidth]{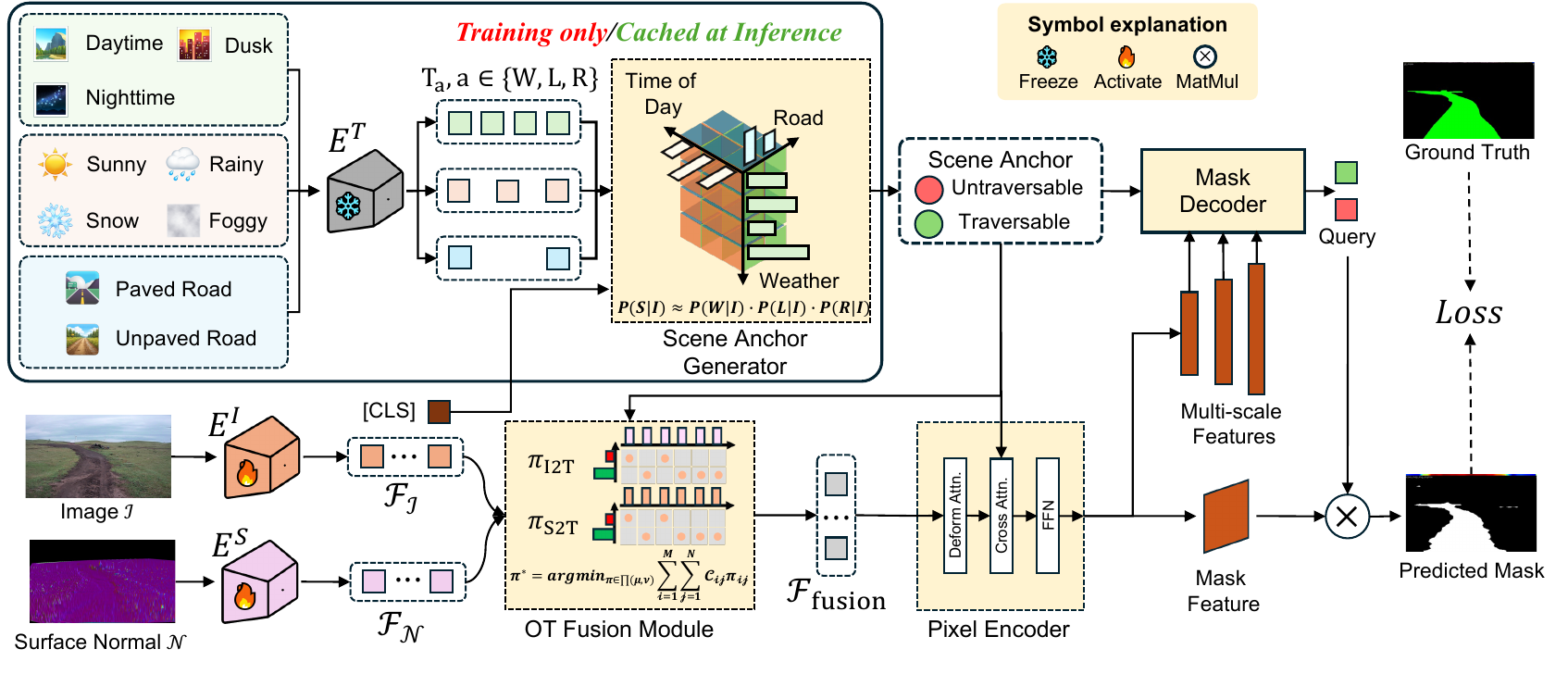}
	\caption{Overall architecture of the proposed OT-Drive. 
	The framework is composed of two modules designed for OOD generalization in traversable area segmentation: 
	1) The Scene Anchor Generator (SAG) constructs scene-specific semantic anchors from the input image. 
	2) The OT Fusion Module leverages optimal transport to align RGB and surface normal features onto the manifold spanned by the semantic anchors, achieving distribution-level feature fusion.}
	\label{fig:framework}
\end{figure*}

\subsection{Problem Formulation}
\label{sec:problem_formulation}
Let $\mathcal{S}$ denote the space of all real-world scene attribute combinations, such as weather (atmospheric conditions), time-of-day (temporal conditions), and road type (terrain properties). However, the training set typically covers only a sparse subset of this space:
\begin{equation}
	\mathcal{S}_{\text{train}} \subset \mathcal{S}, \quad |\mathcal{S}_{\text{train}}| \ll |\mathcal{S}|
\end{equation}

The core challenge of OOD generalization arises because the test set contains novel attribute combinations absent from training:
\begin{equation}
	\mathcal{S}_{\text{test}} \not\subseteq \mathcal{S}_{\text{train}}, \quad \text{i.e., } \exists\, s \in \mathcal{S}_{\text{test}} : s \notin \mathcal{S}_{\text{train}}
\end{equation}

To address this challenge, let $\mathcal{I}$ and $\mathcal{N}$ denote the input spaces for RGB image and surface normal, respectively, and let $\mathcal{Y}$ represent the traversability label space. Our objective is to learn a robust mapping $f: (\mathcal{I}, \mathcal{N}) \rightarrow \mathcal{Y}$ that generalizes to OOD scenes:
\begin{equation}
	\min_{f} \mathbb{E}_{(x,y) \sim \mathcal{D}_{\text{test}}} \left[ \mathcal{L}_{\text{seg}}(f(x), y) \right], \quad \forall\, \mathcal{S}_{\text{test}} \not\subseteq \mathcal{S}_{\text{train}}
\end{equation}
where $x = (I, N)$ denotes the multi-modal input and $\mathcal{L}_{\text{seg}}$ is the segmentation loss.

\subsection{Scene Anchor Generator}
\label{sec:scene_decoupling}

Recent Vision-Language Models (VLMs)~\cite{radfordLearningTransferableVisual2021,jia2021scaling,fangEVA02VisualRepresentation2024} have shown that language semantics are inherently domain-invariant: while visual features vary significantly across domains, the language semantics remain consistent. 
This motivates us to leverage VLMs to construct domain-invariant scene anchors.

\begin{figure}[htbp]
	\centering
	\includegraphics[width=\linewidth]{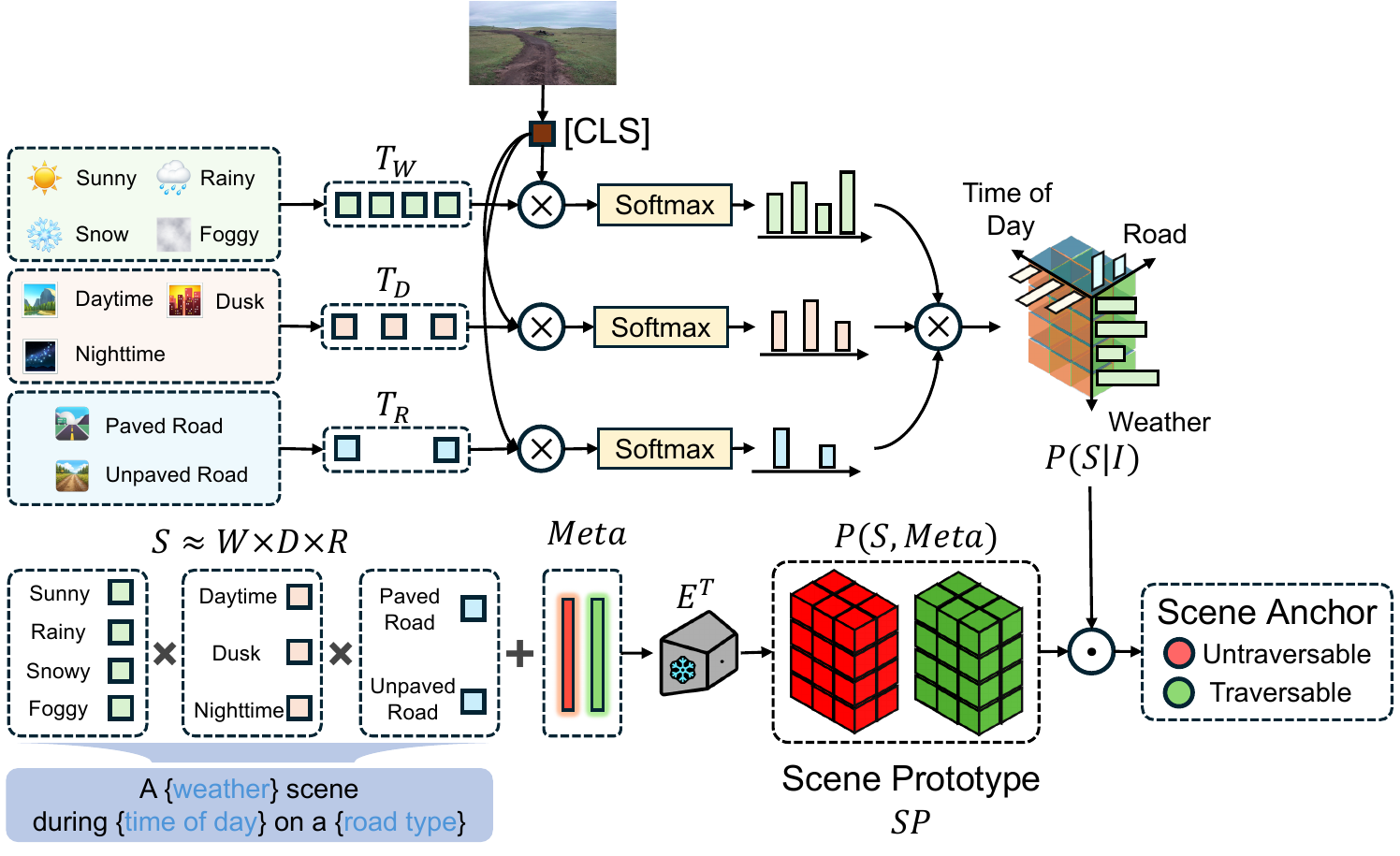}
	\caption{
		Scene Anchor Generator (SAG). 
		Top: Scene attributes (weather, time-of-day, road type) are classified to derive the scene distribution.
		Bottom: Scene prototypes are weighted by the distribution to form the scene anchor.
	}
	\label{fig:scene_decouple}
\end{figure}

\subsubsection{Scene Marginal Probability Modeling}

Our key insight is that rather than modeling $P(\mathcal{S})$ holistically, we can approximate the scene distribution using marginal probabilities of key attributes for real-time efficiency.

In this paper, we select weather (atmospheric conditions), time-of-day (temporal conditions), and road type (terrain properties) as the key attributes, which align with the annotation schema of major off-road datasets~\cite{minORFDDatasetBenchmark2022,minAdvancingOffroadAutonomous2025}.

We factorize the scene distribution into marginal distributions:
\begin{equation}
P(\mathcal{S} | I) \approx P(W | I) \otimes P(D | I) \otimes P(R | I),
\label{eq:scene_factorization}
\end{equation}
where $\otimes$ denotes the tensor product, $W$ denotes the weather, $D$ denotes the time-of-day, and $R$ denotes the road type.

Leveraging the image-text contrastive capability of VLMs, we build three lightweight classifiers using the image \texttt{[CLS]} token. The probability for each attribute $a \in \{W, L, R\}$ is computed as:
\begin{equation}
p(a_i | I) = \frac{\exp(\langle \text{CLS}_{\text{img}}, T_{a_i} \rangle / \tau)}{\sum_j \exp(\langle \text{CLS}_{\text{img}}, T_{a_j} \rangle / \tau)},
\end{equation}
where $T_{a_i}$ denotes the text embedding for the $i$-th category of attribute $a$, and $\tau$ is a learnable temperature parameter.

\subsubsection{Scene Anchor Generation}
As illustrated in the lower part of Fig.~\ref{fig:scene_decouple}, we construct a static Scene Prototype $SP$ by enumerating the combinatorial space of disjoint attributes ($W \times D \times R$). Additionally, we introduce two learnable Meta-Traversability embeddings $Meta \in \mathbb{R}^{2}$ to encode the latent semantics of traversable and non-traversable regions:
\begin{equation}
SP = E_T(\mathcal{S}, Meta),
\end{equation}
where $\mathcal{S}$ denotes the set of all possible scene combinations.

The Scene Anchor $\bar{T}_{S}$ for the current environment is then synthesized by aggregating the prototypes, weighted by their posterior probabilities inferred from the input image:
\begin{equation}
\bar{T}_{S} = \sum_{s \in \mathcal{S}} P(s|I) \cdot E_T(s, Meta),
\end{equation}
where $\mathcal{S}$ denotes the set of all possible scene combinations, and $E_T(\cdot)$ is the frozen text encoder.

\subsubsection{Inference Efficiency}
A key advantage of this design is its efficiency: the text encoder is only used during training, and at deployment we simply weight the cached prototypes $SP$ with inferred posterior probabilities to derive scene anchors.
This achieves zero-overhead semantic guidance by eliminating the text encoder at inference, 
improving speed by 25.6\% (from 2.97 to 3.73 FPS).

\subsection{Multi-modal Fusion via Optimal Transport}
\label{sec:ot_fusion}

According to the manifold hypothesis~\cite{tenenbaum2000global}, features from different modalities lie on distinct low-dimensional manifolds. 
Conventional pixel-level or token-level alignment assumes rigid point-wise correspondence between these manifolds, which breaks down under OOD scenarios. 

Our key insight is that by transporting visual features from different modalities (RGB and surface normal) onto the scene anchor manifold, we can perform feature fusion on a domain-invariant representation space.

\begin{figure}[htbp]
	\centering
	\includegraphics[width=\linewidth]{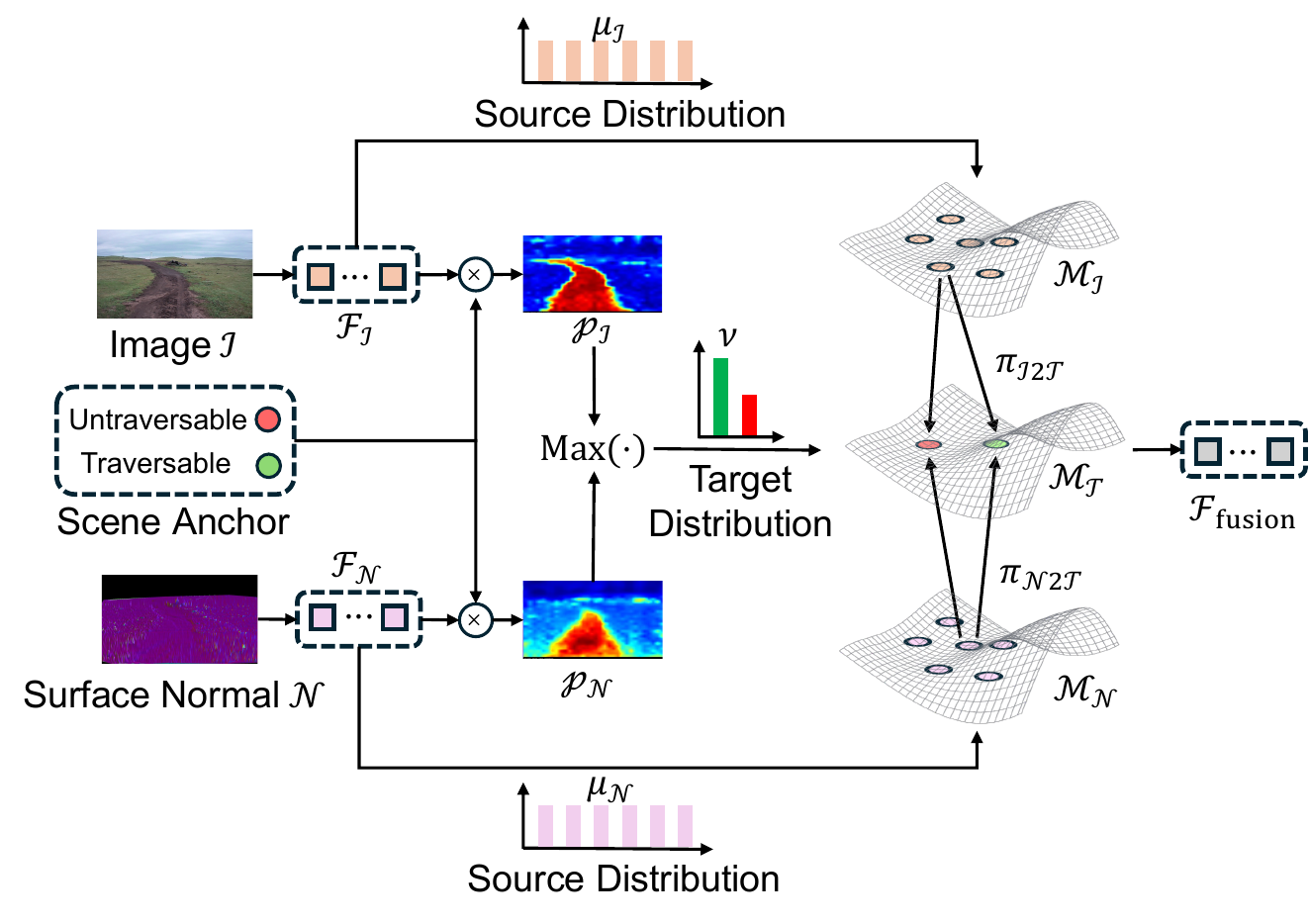}
	\caption{
		OT Fusion Module. 
		The image and normal features are then transported to the scene anchor manifold via optimal transport for fusion.
	}
	\label{fig:optimal_transport}
\end{figure}

\subsubsection{Source Distribution and Target Distribution}

As illustrated in Fig.~\ref{fig:optimal_transport}, we treat the extracted feature maps $F_{\mathcal{I}}, F_{\mathcal{N}} \in \mathbb{R}^{H' \times W' \times C}$ as uniform source distributions over the pixel grid:
\begin{equation}
\mu_{\mathcal{I}} = \frac{1}{N}\sum_{i=1}^{N}\delta_{f_i^{\mathcal{I}}}, \quad \mu_{\mathcal{N}} = \frac{1}{N}\sum_{i=1}^{N}\delta_{f_i^{\mathcal{N}}},
\end{equation}
where $N = H' \times W'$ denotes the number of spatial locations and $\delta$ is the Dirac delta function.

The target distribution $\nu$ is derived by fusing the pre-segmentation probability maps from both modalities via element-wise max pooling:
\begin{equation}
\nu = \sum_{k=1}^{K} m_k \delta_{T_k}, \quad m_k = \frac{\max(P_{\mathcal{I}}^{(k)}, P_{\mathcal{N}}^{(k)})}{\sum_{j=1}^{K} \max(P_{\mathcal{I}}^{(j)}, P_{\mathcal{N}}^{(j)})},
\end{equation}
where $T_k$ denotes the $k$-th semantic anchor and $P_{\mathcal{I}}^{(k)}, P_{\mathcal{N}}^{(k)}$ are the pre-segmentation probabilities for class $k$ from the image and surface normal branches, respectively. Max pooling selects the most confident prediction from either view, ensuring robust target mass estimation.

\subsubsection{Optimal Transport Plan}

Given source distributions $\mu_{\mathcal{I}}$ and $\mu_{\mathcal{N}}$ from the image and surface normal branches, we solve two parallel entropy-regularized OT problems. 

Taking the image branch as an example, we first define the cost matrix $C$ using cosine distance:
\begin{equation}
C_{ik} = 1 - \frac{f_i \cdot T_k}{\|f_i\|_2 \|T_k\|_2}.
\end{equation}
The optimal transport plan is then obtained by minimizing:
\begin{equation}
\pi^*_{\mathcal{I}} = \mathop{\arg\min}_{\pi \in \Pi(\mu_{\mathcal{I}}, \nu)} \langle C_{\mathcal{I}}, \pi \rangle - \varepsilon H(\pi),
\end{equation}
where $\Pi(\mu_{\mathcal{I}}, \nu)$ is the set of joint couplings with marginals $\mu_{\mathcal{I}}$ and $\nu$, $H(\pi)$ is the Shannon entropy, and $\varepsilon$ is the regularization parameter. 

The surface normal branch $\pi^*_{\mathcal{N}}$ is computed analogously. 
Both optimal solutions are obtained efficiently via Sinkhorn-Knopp iterations~\cite{sinkhornConcerningNonnegativeMatrices1967}, which are fully differentiable, enabling end-to-end training.

\subsubsection{Barycentric Projection and Fusion}

We then project both modalities onto the scene anchor $\bar{T}_{S}$ via barycentric mapping and fuse them by averaging:
\begin{align}
\widetilde{F}_{\mathcal{I}} &= \pi^*_{\mathcal{I}} \cdot \bar{T}_{S}, \quad \widetilde{F}_{\mathcal{N}} = \pi^*_{\mathcal{N}} \cdot \bar{T}_{S}, \\
F_{\text{fusion}} &= \lambda \widetilde{F}_{\mathcal{I}} + (1-\lambda) \widetilde{F}_{\mathcal{N}},
\end{align}
where $\lambda$ is the fusion weight. After projection, both modalities reside in the same semantic space, enabling straightforward cross-modal fusion.

\subsection{Masked Decoder}

The fused features $F_{\text{fusion}}$ are decoded into mask features $F_{\text{mask}}$ and multi-scale features $F_{\text{multi}}$. Crucially, we initialize the traversability queries using the Scene Anchor $\bar{T}_S$, which explicitly conditions the segmentation process on scene priors:
\begin{align}
Q_S^{(0)} &= \bar{T}_S + Q_{\text{pos}}, \\
Q_S^{(l)} &= \mathrm{DecoderLayer}\left(Q_S^{(l-1)}, F_{\text{multi}}, M^{(l-1)}\right),
\end{align}
where $Q_{\text{pos}}$ is the learnable positional encoding and $M^{(l-1)}$ is the mask from the previous layer.

The final segmentation mask is produced by computing cosine similarity between refined queries and mask features:
\begin{equation}
M^{(L)} = \sigma\left(\frac{Q_S^{(L)} \cdot F_{\text{mask}}}{\|Q_S^{(L)}\|_2 \|F_{\text{mask}}\|_2 + \epsilon}\right),
\end{equation}
where $\sigma$ denotes the Sigmoid function. The output $M^{(L)}$ serves as the final traversability prediction.

\subsection{Loss Function}

The total training objective combines four components:

\paragraph{Segmentation Loss.}
Following Mask2Former~\cite{chengMaskedattentionMaskTransformer2022}, we apply auxiliary supervision at each decoder layer using classification, binary cross-entropy, and Dice losses:
\begin{equation}
\mathcal{L}_{\text{seg}} = \sum_{l=1}^{L} \left( \lambda_{\text{cls}}\mathcal{L}_{\text{cls}}^l + \lambda_{\text{bce}}\mathcal{L}_{\text{bce}}^l + \lambda_{\text{dice}}\mathcal{L}_{\text{dice}}^l \right),
\end{equation}
where $\lambda_{\text{cls}}, \lambda_{\text{bce}}, \lambda_{\text{dice}}$ are the loss weights for each term.

\paragraph{Vision-Language Regularization.}
To preserve the pre-trained alignment, we regularize both visual and textual representations:
\begin{equation}
\mathcal{L}_{\text{reg}} = \underbrace{\|\text{CLS}_{\text{img}} - \text{CLS}_{\text{img}}^{\text{frozen}}\|_2^2}_{\text{V2V}} + \underbrace{\mathcal{L}_{\text{CE}}(\bar{T}_{S}, \bar{T}_{\text{frozen}})}_{\text{L2L}},
\end{equation}
where $\text{CLS}_{\text{img}}^{\text{frozen}}$ and $\bar{T}_{\text{frozen}}$ denote the feature embeddings extracted by the fixed pre-trained encoders. 

\paragraph{Scene Classification Loss.}
To supervise the SAG, we impose a cross-entropy classification loss $\mathcal{L}_{\text{cls}}^{\text{scene}}$ on the predicted factor distributions. 

\paragraph{Total Objective.}
\begin{equation}
\mathcal{L}_{\text{total}} = \lambda_{\text{1}}\mathcal{L}_{\text{seg}} + \lambda_{\text{2}}\mathcal{L}_{\text{reg}} + \lambda_{\text{3}}\mathcal{L}_{\text{cls}}^{\text{scene}}.
\end{equation}
where $\lambda_{\text{1}}, \lambda_{\text{2}}, \lambda_{\text{3}}$ are the loss weights for each term.

\section{Experiments}
\subsection{Dataset Information}

To evaluate the model’s generalization capability, we conduct comprehensive experiments on two off-road datasets: ORFD \cite{minORFDDatasetBenchmark2022} and ORAD-3D \cite{minAdvancingOffroadAutonomous2025}. 

\begin{figure}[htbp]
    \centering
    \includegraphics[width=\linewidth]{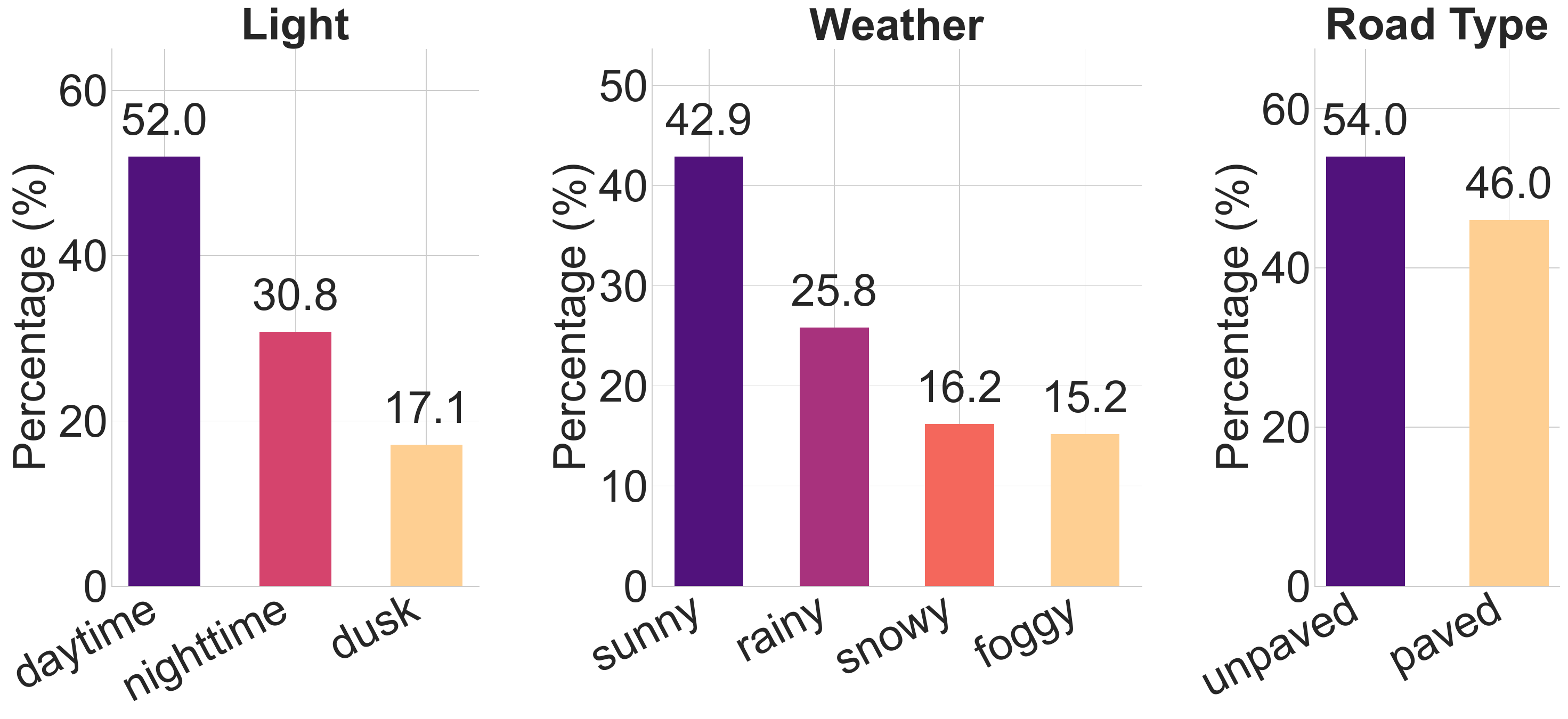}
    \caption{Long-tailed distribution analysis of the ORFD dataset.}
    \label{fig:data_dist}
\end{figure}

Both datasets provide high-resolution images and LiDAR data across diverse weather, time-of-day, and road conditions. The distribution of these attributes is detailed in Fig.~\ref{fig:data_dist}.

\begin{figure}[htbp]
    \centering
    \includegraphics[width=\linewidth]{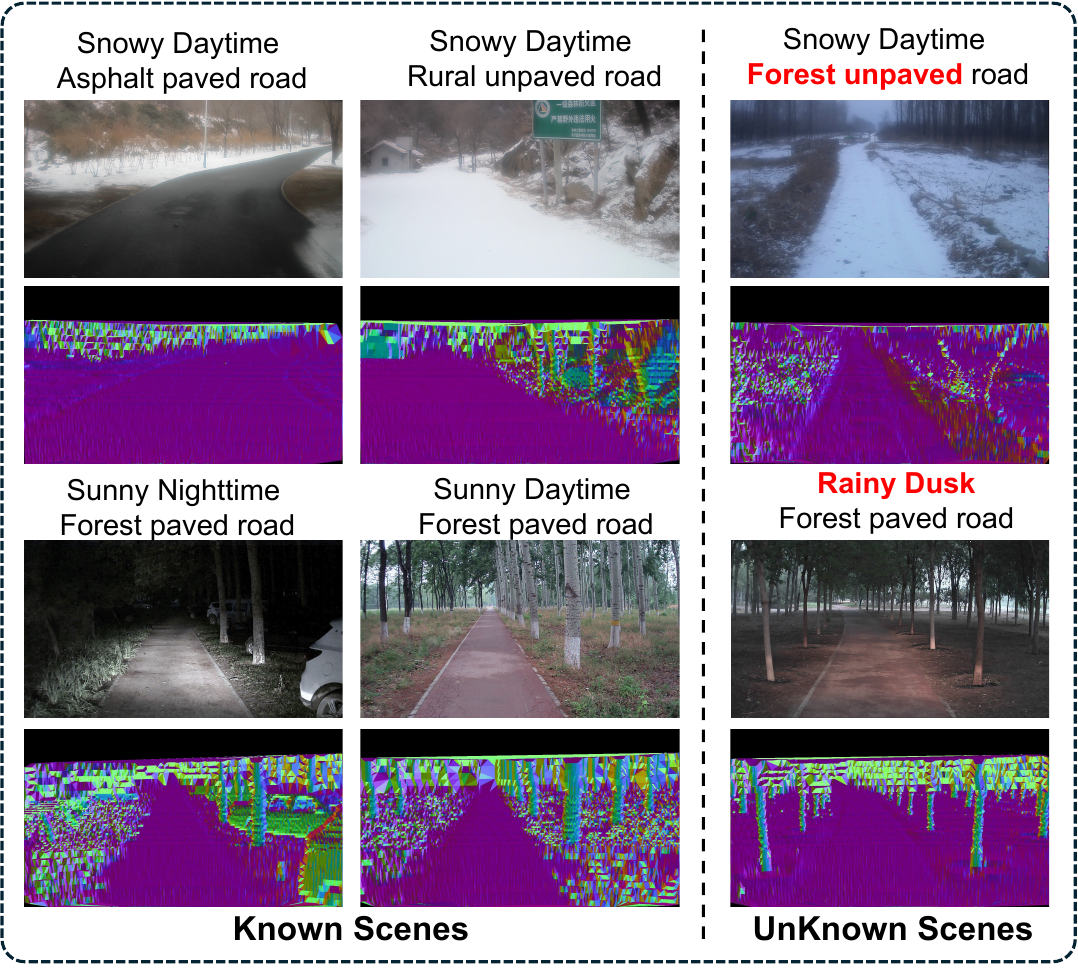}
    \caption{Comparison of Known and Unknown scenes. Red text indicates attribute differences from the training distribution.}
    \label{fig:scene_cls}
\end{figure}

Following Sec.~\ref{sec:problem_formulation}, we partition both datasets into Known (ID) and Unknown (OOD) splits based on scene attribute combinations, where OOD contains only combinations absent from training (see Fig.~\ref{fig:scene_cls}).

Detailed statistics of this splitting protocol are summarized in Table~\ref{tab:orfd_orad_stats}.

\begin{table*}[htbp]
	\centering
	\caption{Comparison of statistical information for the ORFD and ORAD-3D datasets}
	\label{tab:orfd_orad_stats}
	\begin{tabular}{@{}lccccccc@{}}
		\toprule
		\multirow{2}{*}{\textbf{Dataset}} & \multirow{2}{*}{\textbf{Total Samples}} & \multirow{2}{*}{\textbf{Scenes}} & \multirow{2}{*}{\textbf{Split Ratio}} & \multirow{2}{*}{\textbf{Train Set}} & \multirow{2}{*}{\textbf{Val. Set}} & \multicolumn{2}{c}{\textbf{Test Set}} \\
		 \cmidrule(lr){7-8}
		& & & & & & \textbf{Known Scenes} & \textbf{Unknown Scenes} \\ 
		\midrule
		\textbf{ORFD}~\cite{minORFDDatasetBenchmark2022} & 12K & 30 & 7:1:2 & 8.3K (21) & 1.6K (4) & 1.4K (3) & 0.7K(2) \\[2pt]
		\textbf{ORAD-3D}~\cite{minAdvancingOffroadAutonomous2025} & 57K & 142 & 7:1:2 & 40K (102) & 5.7K (15) & 9.2K (22) & 1.9K(4) \\
		\textbf{ORFD-to-ORAD3D} & 11.1K & 26 & - & - & - & 9.3K (22) & 1.8K(4) \\ 
		\bottomrule
	\end{tabular}
	\footnotesize
\end{table*}

\begin{table*}[!htbp]
	\centering
	\caption{Performance comparison on the ORFD test set}
	\label{tab:comparison_results_orfd}
	\resizebox{\textwidth}{!}{%
		\begin{tabular}{l|cccc|cccc|cccc|cccc}
			\toprule
			\multirow{2}{*}{Model} & \multicolumn{4}{c|}{Overall Metrics (\%)} & \multicolumn{4}{c|}{Known Scenes (\%)} & \multicolumn{4}{c|}{Unknown Scenes (\%)} & \multicolumn{4}{c}{$\Delta$ (Unknown - Known) (\%)} \\
			\cmidrule(lr){2-17}
			& mAcc & mRecall & mF1 & mIoU & mAcc & mRecall & mF1 & mIoU & mAcc & mRecall & mF1 & mIoU & mAcc & mRecall & mF1 & mIoU \\
			\midrule
			OFF-Net~\cite{minORFDDatasetBenchmark2022}   & 89.07 & 93.65 & 90.87 & 83.48 & 95.53 & 96.39 & 95.95 & 92.25 & 78.08 & 89.47 & 80.62 & 68.42 & \cellcolor{red!10}\textcolor{red}{$\downarrow$17.45} & \cellcolor{red!10}\textcolor{red}{$\downarrow$6.92} & \cellcolor{red!10}\textcolor{red}{$\downarrow$15.33} & \cellcolor{red!10}\textcolor{red}{$\downarrow$23.83} \\
			M2F2-Net~\cite{yeM2F2netMultimodalFeature2023}  & 88.80 & 93.92 & 90.73 & 83.23 & 94.58 & \underline{96.47} & 95.43 & 91.31 & 78.44 & 89.88 & 81.05 & 69.01 & \cellcolor{red!10}\textcolor{red}{$\downarrow$16.14} & \cellcolor{red!10}\textcolor{red}{$\downarrow$6.59} & \cellcolor{red!10}\textcolor{red}{$\downarrow$14.38} & \cellcolor{red!10}\textcolor{red}{$\downarrow$22.30} \\
			ROD~\cite{sunRODRGBonlyFast2025} & \underline{97.14} & \underline{95.02} & \underline{96.01} & \underline{92.40} & \underline{97.43} & 96.13 & \underline{96.75} & \underline{93.73} & \underline{96.51} & \underline{91.83} & \underline{93.94} & \underline{88.81} & \cellcolor{red!10}\textcolor{red}{$\downarrow$0.92} & \cellcolor{red!10}\textcolor{red}{$\downarrow$4.30} & \cellcolor{red!10}\textcolor{red}{$\downarrow$2.81} & \cellcolor{red!10}\textcolor{red}{$\downarrow$4.92} \\
			\midrule
			OT-Drive (Ours)       & \textbf{97.70} & \textbf{97.57} & \textbf{97.63} & \textbf{95.40} & \textbf{97.78} & \textbf{97.53} & \textbf{97.65} & \textbf{95.42} & \textbf{97.34} & \textbf{97.66} & \textbf{97.50} & \textbf{95.16} & \cellcolor{green!15}\textcolor{green!60!black}{\textbf{$\downarrow$0.44}} & \cellcolor{green!15}\textcolor{green!60!black}{\textbf{$\uparrow$0.13}} & \cellcolor{green!15}\textcolor{green!60!black}{\textbf{$\downarrow$0.15}} & \cellcolor{green!15}\textcolor{green!60!black}{\textbf{$\downarrow$0.26}} \\
			\bottomrule
		\end{tabular}%
	}
\end{table*}

\begin{table*}[htbp]
	\centering
	\caption{Performance comparison on the ORAD-3D test set}
	\label{tab:comparison_results_orad}
	\resizebox{\textwidth}{!}{%
		\begin{tabular}{l|cccc|cccc|cccc|cccc}
			\toprule
			\multirow{2}{*}{Model} & \multicolumn{4}{c|}{Overall Metrics (\%)} & \multicolumn{4}{c|}{Known Scenes (\%)} & \multicolumn{4}{c|}{Unknown Scenes (\%)} & \multicolumn{4}{c}{$\Delta$ (Unknown - Known) (\%)} \\
			\cmidrule(lr){2-17}
			& mAcc & mRecall & mF1 & mIoU & mAcc & mRecall & mF1 & mIoU & mAcc & mRecall & mF1 & mIoU & mAcc & mRecall & mF1 & mIoU \\
			\midrule
			OFF-Net~\cite{minORFDDatasetBenchmark2022}   & 93.92 & 93.07 & 93.48 & 87.86 & 94.12 & 93.35 & 93.72 & 88.29 & \underline{92.94} & 91.67 & 92.27 & 85.83 & \cellcolor{red!10}\textcolor{red}{$\downarrow$1.18} & \cellcolor{red!10}\textcolor{red}{$\downarrow$1.68} & \cellcolor{red!10}\textcolor{red}{$\downarrow$1.45} & \cellcolor{red!10}\textcolor{red}{$\downarrow$2.46} \\
			M2F2-Net~\cite{yeM2F2netMultimodalFeature2023}  & \underline{94.58} & \underline{93.80} & \underline{94.18} & \underline{89.09} & \underline{94.92} & \underline{94.05} & \underline{94.47} & \underline{89.60} & \underline{92.94} & \underline{92.61} & \underline{92.77} & \underline{86.67} & \cellcolor{red!10}\textcolor{red}{$\downarrow$1.98} & \cellcolor{red!10}\textcolor{red}{$\downarrow$1.44} & \cellcolor{red!10}\textcolor{red}{$\downarrow$1.70} & \cellcolor{red!10}\textcolor{red}{$\downarrow$2.93} \\
			ROD~\cite{sunRODRGBonlyFast2025} & 91.71 & 83.08 & 85.92 & 75.92 & 92.29 & 84.11 & 86.86 & 77.29 & 88.80 & 78.00 & 81.14 & 69.42 & \cellcolor{red!10}\textcolor{red}{$\downarrow$3.49} & \cellcolor{red!10}\textcolor{red}{$\downarrow$6.11} & \cellcolor{red!10}\textcolor{red}{$\downarrow$5.72} & \cellcolor{red!10}\textcolor{red}{$\downarrow$7.87} \\
			\midrule
			OT-Drive (Ours)       & \textbf{95.32} & \textbf{94.92} & \textbf{95.12} & \textbf{90.75} & \textbf{95.41} & \textbf{95.03} & \textbf{95.22} & \textbf{90.93} & \textbf{94.91} & \textbf{94.35} & \textbf{94.63} & \textbf{89.89} & \cellcolor{green!15}\textcolor{green!60!black}{\textbf{$\downarrow$0.50}} & \cellcolor{green!15}\textcolor{green!60!black}{\textbf{$\downarrow$0.68}} & \cellcolor{green!15}\textcolor{green!60!black}{\textbf{$\downarrow$0.59}} & \cellcolor{green!15}\textcolor{green!60!black}{\textbf{$\downarrow$1.04}} \\
			\bottomrule
		\end{tabular}%
	}
\end{table*}

\subsection{Experimental Setup}

We employ EVA02-CLIP-B-16~\cite{fangEVA02VisualRepresentation2024} as the visual backbone at $512\times512$ resolution. Surface normals are estimated using D2NT~\cite{fengD2NTHighPerformingDepthtoNormal2023}. The decoder follows a two-stage design: a Pixel Decoder~\cite{pakTextualQuerydrivenMask2025} with multi-scale deformable attention. The model is trained for 20,000 iterations with a batch size of 16 using 
AdamW optimizer (learning rate $1\text{e}{-5}$, weight decay $1\text{e}{-4}$). 
The fusion weight $\lambda$ in Eq.~(12) is set to 0.5.

We report standard semantic segmentation metrics: mean Accuracy (mAcc), mean Recall (mRec), mean F1-score (mF1), and mean Intersection over Union (mIoU). 

\subsection{Baselines}
\begin{itemize}
	\item OFF-Net~\cite{minORFDDatasetBenchmark2022}: A Transformer-based fusion network combining image and surface normal.
	\item M2F2-Net~\cite{yeM2F2netMultimodalFeature2023}: Features multi-modal cross-fusion and edge-aware decoding for improved boundary accuracy.
	\item ROD~\cite{sunRODRGBonlyFast2025}: A single-modal RGB method using pretrained ViT for efficient free-space detection.
\end{itemize}
\subsection{Quantitative Analysis}

To evaluate the generalization capability of OT-Drive, we conduct extensive comparative evaluations against three baselines under two distinct settings:

\subsubsection{Within-Dataset Performance}
\begin{figure}[htbp]
	\centering
	\includegraphics[width=\linewidth]{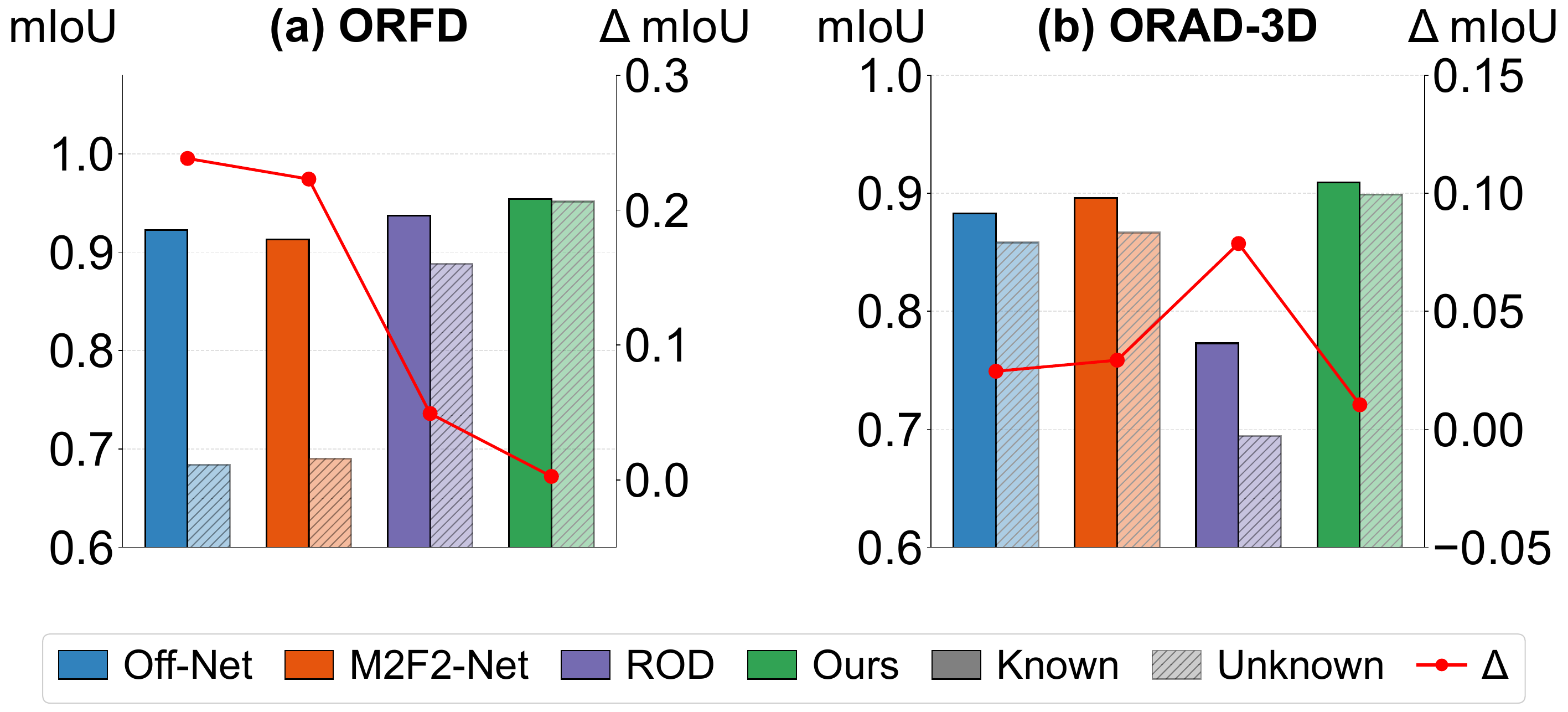}
	\caption{Relative performance degradation ($\Delta$mIoU, \%) from known to unknown scenes. Lower values indicate stronger generalization capability.}
	\label{fig:exp}
\end{figure}

Table~\ref{tab:comparison_results_orfd}, Table~\ref{tab:comparison_results_orad}, and Fig.~\ref{fig:exp} present quantitative results on ORFD and ORAD-3D. 

\begin{enumerate}
	\item \textbf{Generalization to OOD Scenarios:} 
	Our method achieves minimal performance degradation in OOD scenarios ($\Delta mIoU < 1.1\%$), substantially outperforming baselines (see \figurename{\ref{fig:exp}}).
	Existing methods rely on memorized dataset-specific patterns that fail to generalize to novel scenes. Our SAG addresses this by constructing domain-invariant scene anchors from language semantics, effectively enhancing OOD generalization.

    \item \textbf{Robustness of OT Fusion:} 
	Our method achieves the best performance on both datasets, consistently outperforming existing methods in both ID and OOD scenarios (see Tables~\ref{tab:comparison_results_orfd}--\ref{tab:comparison_results_orad}).
	Unlike conventional methods that rely on pixel/token-level alignment sensitive to domain variations, our OT Fusion aligns features at the distribution level via transport constraints, yielding robust predictions across all scenarios.
\end{enumerate}

\subsubsection{Cross-Dataset Generalization}
As presented in Table~\ref{tab:cross_dataset}, we perform a cross-dataset transfer evaluation by training on ORFD (8.3K) and testing directly on ORAD-3D (11.1K) without fine-tuning. 

Our method demonstrates strong data efficiency: even when tested on a larger dataset (11.1K) than training (8.3K), it still achieves the best OOD generalization (outperforming baselines by 13.99\%) with minimal performance degradation ($\Delta mIoU = +0.07\%$).
This is because we do not directly model the entire scene distribution, but instead factorize it into marginal probabilities over independent attributes, enabling compositional generalization by recombining scene factors learned from known scenarios.

\begin{table*}[htbp]
	\centering
	\caption{Cross-dataset generalization: trained on ORFD, tested on ORAD-3D test set (26 scenes)}
	\label{tab:cross_dataset}
	\resizebox{\textwidth}{!}{%
		\begin{tabular}{l|cccc|cccc|cccc|cccc}
			\toprule
			\multirow{2}{*}{Model} & \multicolumn{4}{c|}{Overall Metrics (\%)} & \multicolumn{4}{c|}{Known Scenes (\%)} & \multicolumn{4}{c|}{Unknown Scenes (\%)} & \multicolumn{4}{c}{$\Delta$ (Unknown - Known) (\%)} \\
			\cmidrule(lr){2-17}
			& mAcc & mRecall & mF1 & mIoU & mAcc & mRecall & mF1 & mIoU & mAcc & mRecall & mF1 & mIoU & mAcc & mRecall & mF1 & mIoU \\
			\midrule
			OFF-Net~\cite{minORFDDatasetBenchmark2022}   & 92.23 & 92.05 & 92.13 & 85.57 & \underline{93.98} & 93.43 & \underline{93.70} & \underline{88.24} & 83.18 & 84.70 & 83.88 & 72.85 & \cellcolor{red!10}\textcolor{red}{$\downarrow$10.80} & \cellcolor{red!10}\textcolor{red}{$\downarrow$8.73} & \cellcolor{red!10}\textcolor{red}{$\downarrow$9.82} & \cellcolor{red!10}\textcolor{red}{$\downarrow$15.39} \\
			M2F2-Net~\cite{yeM2F2netMultimodalFeature2023}  & 91.44 & \underline{93.45} & \underline{92.33} & \underline{85.87} & 93.00 & \textbf{94.24} & 93.58 & 88.02 & 84.23 & \underline{89.69} & \underline{86.00} & \underline{75.80} & \cellcolor{red!10}\textcolor{red}{$\downarrow$8.77} & \cellcolor{red!10}\textcolor{red}{$\downarrow$4.55} & \cellcolor{red!10}\textcolor{red}{$\downarrow$7.58} & \cellcolor{red!10}\textcolor{red}{$\downarrow$12.22} \\
			ROD~\cite{sunRODRGBonlyFast2025} & \underline{92.28} & 86.42 & 88.64 & 79.97 & 92.33 & 87.14 & 89.14 & 80.73 & \underline{92.32} & 82.21 & 85.62 & 75.65 & \cellcolor{red!10}\textcolor{red}{$\downarrow$0.01} & \cellcolor{red!10}\textcolor{red}{$\downarrow$4.93} & \cellcolor{red!10}\textcolor{red}{$\downarrow$3.52} & \cellcolor{red!10}\textcolor{red}{$\downarrow$5.08} \\
			\midrule
			OT-Drive (Ours)       & \textbf{95.03} & \textbf{94.10} & \textbf{94.55} & \textbf{89.74} & \textbf{95.01} & \underline{94.11} & \textbf{94.54} & \textbf{89.72} & \textbf{95.14} & \textbf{94.02} & \textbf{94.56} & \textbf{89.79} & \cellcolor{green!15}\textcolor{green!60!black}{\textbf{$\uparrow$0.13}} & \cellcolor{green!15}\textcolor{green!60!black}{\textbf{$\downarrow$0.09}} & \cellcolor{green!15}\textcolor{green!60!black}{\textbf{$\uparrow$0.02}} & \cellcolor{green!15}\textcolor{green!60!black}{\textbf{$\uparrow$0.07}} \\
			\bottomrule
		\end{tabular}%
	}
\end{table*}

\subsection{Ablation Study}

To validate the contributions of each proposed component, we conduct progressive ablation experiments, with results summarized in Table~\ref{tab:ablation}.

\begin{table*}[t]
	\centering
	\caption{Ablation study on ORFD and ORAD-3D datasets.}
	\label{tab:ablation}
	\resizebox{0.85\textwidth}{!}{
	\begin{tabular}{c|ccc|cccc|cccc}
	\toprule
	\multirow{2}{*}{No.} & \multicolumn{3}{c|}{Components} & \multicolumn{4}{c|}{ORFD (mIoU \%)} & \multicolumn{4}{c}{ORAD-3D (mIoU \%)} \\
	\cmidrule(lr){2-12}
	& Scene Anchor & OT Fusion & Marginal Prob. & Overall & Known & Unknown & $\Delta$ & Overall & Known & Unknown & $\Delta$ \\
	\midrule
	1 & \phantom{x} & \phantom{x} & \phantom{x} & 85.44 & 92.14 & 73.06 & \cellcolor{red!10}\textcolor{red}{$\downarrow$19.08} & 87.96 & 88.89 & 83.53 & \cellcolor{red!10}\textcolor{red}{$\downarrow$5.35} \\
	2 & \checkmark & \phantom{x} & \phantom{x} & 94.35 & 94.36 & 94.09 & \cellcolor{red!10}\textcolor{red}{$\downarrow$0.27} & 90.57 & 90.85 & 89.26 & \cellcolor{red!10}\textcolor{red}{$\downarrow$1.58} \\
	3 & \checkmark & \checkmark & \phantom{x} & \textbf{95.48} & \textbf{95.98} & \underline{94.15} & \cellcolor{red!10}\textcolor{red}{$\downarrow$1.83} & 90.55 & 90.73 & \underline{89.65} & \cellcolor{red!10}\textcolor{red}{$\downarrow$1.08} \\
	\midrule
	Ours & \checkmark & \checkmark & \checkmark & \underline{95.40} & \underline{95.42} & \textbf{95.16} & \cellcolor{green!10}\textcolor{green!60!black}{$\textbf{$\downarrow$0.26}$} & \textbf{90.75} & \textbf{90.93} & \textbf{89.89} & \cellcolor{green!10}\textcolor{green!60!black}{$\textbf{$\downarrow$1.04}$} \\
	\bottomrule
	\end{tabular}
	}
\end{table*}

\begin{enumerate}
    \item \textbf{Effectiveness of Scene Anchor.} 
	Comparing Model \#1 with Model \#2 demonstrates substantial OOD generalization improvements on both ORFD and ORAD-3D, with mIoU gains of 21.03\% and 5.73\%, respectively. This validates that scene anchors effectively enhance the model's generalization capability.
    
    \item \textbf{Effectiveness of OT Fusion.} 
	Comparing Model \#2 with Model \#3 reveals consistent performance gains in OOD scenarios, validating that distribution-level feature fusion via optimal transport can achieve robust generalization when encountering OOD features in unseen environments.
    
    \item \textbf{Effectiveness of Marginal Probability Modeling.} 
	Comparing Model \#3 with the full model demonstrates that marginal probability modeling is essential for generalizing to unseen scene combinations, as it enables factor recombination and significantly narrows the generalization gap on both datasets.
	
\end{enumerate}

\subsection{Qualitative Analysis}
\begin{figure}[htbp]
	\centering
	\includegraphics[width=\linewidth]{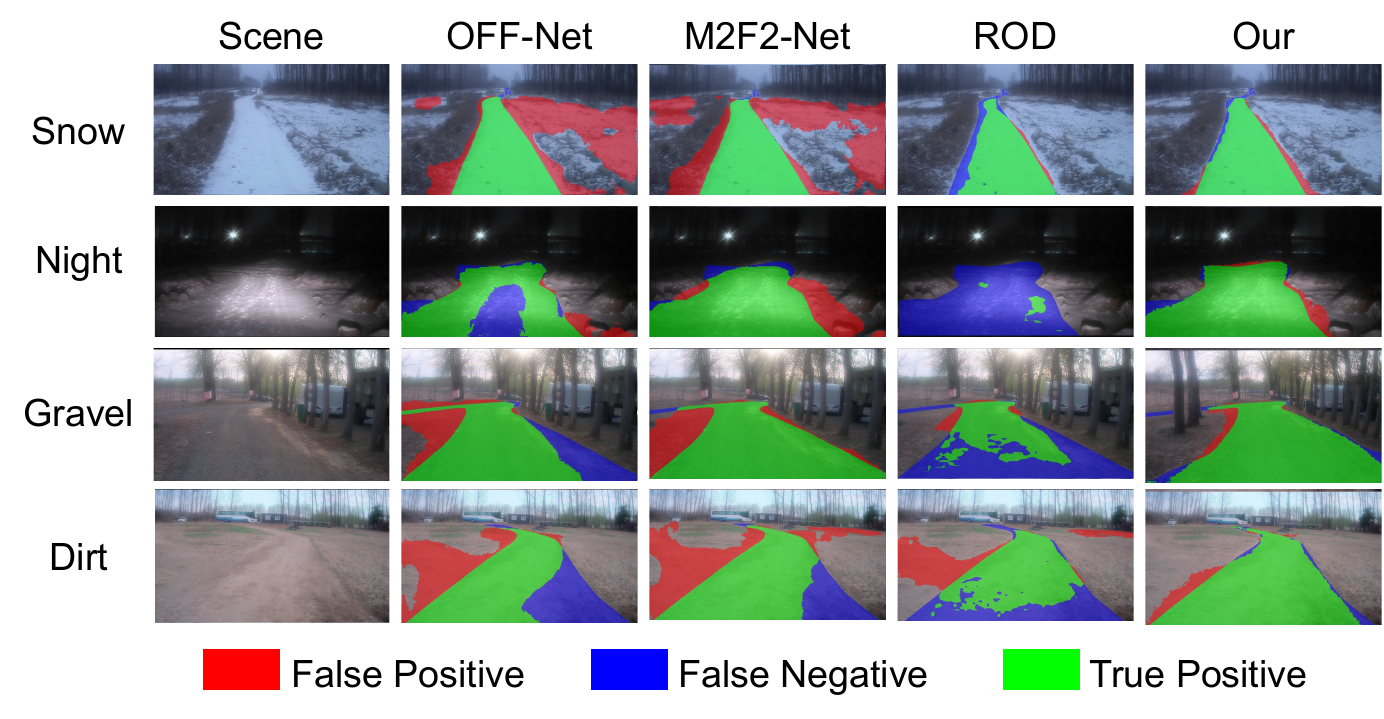}
	\caption{Qualitative comparison of segmentation results. From left to right: RGB input, OFF-Net~\cite{minORFDDatasetBenchmark2022}, M2F2-Net~\cite{yeM2F2netMultimodalFeature2023}, ROD~\cite{sunRODRGBonlyFast2025}, and OT-Drive (Ours).}
	\label{fig:qualitative}
\end{figure}
We present qualitative comparisons between our method and baseline models in Fig.~\ref{fig:qualitative}, specifically targeting unknown scene combinations.

Existing methods show inconsistent performance: multi-modal methods excel at night but struggle on snow, while RGB-only models exhibit the opposite behavior.
However, our method exhibits consistent superiority across both conditions, validating the effectiveness of our distribution-level OT Fusion.
In gravel and dirt road scenarios, existing methods frequently misclassify geometrically flat regions. 
This occurs because their traversability semantics are biased by flat areas in the training set. 
Our SAG addresses this by compositionally combining marginal probability distributions to construct domain-invariant scene anchors that generalize to unseen scenarios.

\subsection{Efficiency Analysis}

To assess practical deployment potential, we benchmark all methods on the NVIDIA Tesla T4---a widely adopted edge computing GPU. 
For multi-modal methods, we include GPU-accelerated surface normal estimation in the total pipeline time.

As shown in Table~\ref{tab:efficiency}, by simply applying standard inference optimizations (PyTorch 2.0 graph compilation and FP16 mixed-precision), OT-Drive achieves 21.11 FPS---a 5.66$\times$ speedup over the base configuration. 
These results demonstrate the strong real-time deployment potential of OT-Drive on resource-constrained edge devices.
\begin{table}[!htbp]
	\centering
	\caption{
		Comparison of inference speed (FPS) on Tesla T4.
		$\dagger$ includes surface normal estimation latency.
		}
	\label{tab:efficiency}
	\resizebox{\linewidth}{!}{
		\begin{tabular}{l|ccc|cc}
			\toprule
			\multirow{2}{*}{Method} & \multirow{2}{*}{OFF-Net$^\dagger$} & \multirow{2}{*}{M2F2-Net$^\dagger$} & \multirow{2}{*}{ROD} & \multicolumn{2}{c}{OT-Drive$^\dagger$ (Ours)} \\
			\cmidrule(lr){5-6}
			& & & & Base & Optimized \\
			\midrule
			FPS & 4.49 & 4.23 & 3.58 & 3.73 & \textbf{21.11} \\
			\bottomrule
		\end{tabular}
	}
\end{table}

\section{Conclusion}

In this paper, we presented OT-Drive, a novel multi-modal fusion framework designed to address the challenge of OOD generalization in off-road traversable area segmentation. 
The Scene Anchor Generator models marginal probabilities of scene attributes and constructs generalizable scene anchors by composing environmental attributes.
The OT Fusion module then aligns RGB and surface normal features onto the semantic manifold defined by these anchors, achieving distribution-level fusion via Optimal Transport.
Extensive experiments demonstrate that OT-Drive significantly outperforms existing baselines, particularly in OOD scenarios, and achieves a substantial 13.99\% mIoU gain in cross-dataset transfer tasks.
Moreover, OT-Drive attains a real-time inference speed of 21.11 FPS, ensuring its suitability for practical deployment. 
We believe this work represents a meaningful exploration of leveraging Vision-Language Models (VLMs) for traversable area segmentation.

Despite these advancements, our method relies on a predefined closed-set scene space to ensure real-time efficiency.
In the future, we aim to extend this framework by exploring the integration of the complex reasoning capabilities and open-world knowledge of Multimodal Large Language Models (MLLMs), enabling more fine-grained and context-aware traversable area segmentation in complex environments.

\ifCLASSOPTIONcaptionsoff
  \newpage
\fi

\bibliographystyle{IEEEtran}
\bibliography{main.bib}

\end{document}